\documentclass{article}

\usepackage{arxiv}

\usepackage[utf8]{inputenc} 
\usepackage[T1]{fontenc}    
\usepackage{url}            
\usepackage{booktabs}       
\usepackage{amsfonts}       
\usepackage{nicefrac}       
\usepackage{microtype}      
\usepackage{graphicx}
\usepackage[numbers]{natbib}
\usepackage{doi}

\usepackage{multirow}%
\usepackage{comment}

\usepackage{amsmath,amssymb,amsfonts}%
\usepackage{amsthm}%
\usepackage{mathrsfs}%
\usepackage{newtxtext,newtxmath}  
\usepackage{lineno,hyperref}
\usepackage[title]{appendix}%
\usepackage{xcolor}%
\usepackage{textcomp}%
\usepackage{manyfoot}%
\usepackage{longtable}
\usepackage{bm}
\usepackage{array}
\usepackage[caption=false]{subfig}
\usepackage{color}
\usepackage{algorithm}%
\usepackage{algorithmicx}%
\usepackage{algpseudocode}%
\usepackage{listings}%

\usepackage{acro}
\DeclareAcronym{ai}{
    short = AI,
    long = Artificial Intelligence
}

\DeclareAcronym{xai}{
    short = XAI,
    long = eXplainable Artificial Intelligence
}

\DeclareAcronym{darpa}{
    short = DARPA,
    long =  Defense Advanced Research Projects Agency
}

\DeclareAcronym{ml}{
    short = ML,
    long = Machine Learning
}
\providecommand{\keywords}[1]{\textbf{\textit{Keywords---}} #1}

\title{The Role of XAI in Transforming Aeronautics and Aerospace Systems}
\author{
        \href{https://orcid.org/0009-0003-3630-5324}{\includegraphics[scale=0.06]{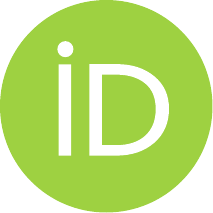}\hspace{1mm}Francisco Javier Cantero} \\
        ENIA IA3 Chair \\
        Universidad de Alcal\'a\\
        Madrid, Spain\\
        \texttt{fco.cantero@uah.es} \\
    \And
        \href{https://orcid.org/0009-0002-9925-5144}{\includegraphics[scale=0.06]{orcid.pdf}\hspace{1mm}Mikel Galafate} \\
        ENIA IA3 Chair \\
        Universidad de Alcal\'a\\
        Madrid, Spain\\
        \texttt{mikel.galafate@uah.es} \\
    \And
        \href{https://orcid.org/0000-0003-1415-1961}{\includegraphics[scale=0.06]{orcid.pdf}\hspace{1mm}Javier M. Moguerza} \\
        CETINIA-Data Science Lab, Universidad Rey Juan Carlos \\
        ENIA IA3 Chair \\
        \texttt{javier.moguerza@urjc.es} \\
    \And
        \href{https://orcid.org/0000-0001-5197-2932}{\includegraphics[scale=0.06]{orcid.pdf}\hspace{1mm}Isaac Mart\'in de Diego} \\
        CETINIA-Data Science Lab, Universidad Rey Juan Carlos \\
        ENIA IA3 Chair \\
        \texttt{isaac.martin@urjc.es} \\        
    \And
        \href{https://orcid.org/0000-0002-9725-0588}{\includegraphics[scale=0.06]{orcid.pdf}\hspace{1mm}M. Teresa Gonz\'alez De Lena} \\
        CETINIA-Data Science Lab, Universidad Rey Juan Carlos \\
        ENIA IA3 Chair \\
        \texttt{mariateresa.gonzalezdelena@urjc.es} \\
    \And
        \href{https://orcid.org/0000-0002-1178-8257}{\includegraphics[scale=0.06]{orcid.pdf}\hspace{1mm} Gema Gutierrez Peña} \\
        CETINIA-Data Science Lab, Universidad Rey Juan Carlos \\
        ENIA IA3 Chair \\
        \texttt{gema.gutierrez@urjc.es} \\
}

\begin{document}
\maketitle

\begin{abstract}
Recent advancements in \ac{ai} have transformed decision-making in aeronautics and aerospace. These advancements in \ac{ai} have brought with them the need to understand the reasons behind the predictions generated by \ac{ai} systems and models, particularly by professionals in these sectors. In this context, the emergence of \ac{xai} has helped bridge the gap between professionals in the aeronautical and aerospace sectors and the \ac{ai} systems and models they work with. For this reason, this paper provides a review of the concept of \ac{xai} is carried out defining the term and the objectives it aims to achieve. Additionally, the paper discusses the types of models defined within it and the properties these models must fulfill to be considered transparent, as well as the post-hoc techniques used to understand \ac{ai} systems and models after their training. Finally, various application areas within the aeronautical and aerospace sectors will be presented, highlighting how \ac{xai} is used in these fields to help professionals understand the functioning of \ac{ai} systems and models.
\end{abstract}


\section{Introduction}\label{section1}
The rise and development of \ac{ai} and its adoption across various sectors of society has led to big advances in decision making in the governments, administrations and businesses. This development of \ac{ai} has led to greater awareness among humans that many of the decisions affecting their daily lives are based on the use of these algorithms or directly taken by these. Due to the relevance \ac{ai} systems and models have taken in our lives, the necessity of understanding the motives of the decisions taken by these models has been spread among researchers and professionals \cite{goodman2016}. This needs to understand the reason behind the decisions comes from the difficulty of understanding the internal workings of an \ac{ai} system or, in other words, when trying to understand the algorithm used by a specific model \cite{castelvecchi2016}. In response to this situation, in 2016, the \ac{darpa}, started the \ac{xai} program, whose objective was to develop several methods that allowed to understand and trust in \ac{ai} systems. As a result, and with the objective of obtaining interpretable and transparent models, \ac{xai} was developed. \ac{xai} proposes the creation of a set of \ac{ml} (ML) techniques that allow the creation of more explainable models maintaining a high training level to have users understand and handle \ac{ai} \cite{gunning2019xai}. 

In aeronautics and aerospace sectors, the use of \ac{ai} systems and models has also brought about significant advancements and evolution of automation and development of tasks and processes in both fields. Along with the development of these industries, the need to know and explain the reasons for the decisions taken by \ac{ai} systems has also risen. Consequently, during the last decade, \ac{xai} arose as a fundamental actor when developing \ac{ai} systems and models. Aerial Traffic Management (ATM) to control aerial traffic \cite{degas2022survey}, or the certification of critical systems that allow the evaluation of the protocols airplane pilots must carry out in critical situations \cite{sutthithatip2022explainable} are some examples of \ac{xai} in aeronautics. In aerospace, \ac{xai} runs a fundamental role in tasks such as generation of explanations for Deep Neural Networks (DNN) used in predictive maintenance, digital twins \cite{shukla2020opportunities}, and the anomaly detection in spacecraft telemetry \cite{cuellar2024explainable}. 

The emergence of \ac{xai} has promoted the use of \ac{ai} in many complex tasks and processes with the goal of facilitating the understanding of \ac{ai} systems and models. For this reason, this article defines \ac{xai} and its objectives in Section \ref{section2}. In Section \ref{section3}, the properties \ac{ai} models must satisfy for them to be considered interpretable and transparent are defined. In Section \ref{section4}, transparent models are defined as well as the techniques applied to models that are not. Section \ref{section5} performs a review on the state-of-the-art of \ac{xai} in aeronautics and aerospace, to end up with the conclusions of the work done in this article in Section \ref{section6}. 

\section{Definition and objectives of \ac{xai}}\label{section2}
The name \ac{xai} was first used when in 2016 \ac{darpa} promoted the \ac{xai} program. The objective of the program was to develop \ac{ml} techniques that allow the creation of more explainable models while maintaining a high-performance level to have users understand and handle \ac{ai} \cite{gunning2019xai}. Since then, numerous researchers and professionals have been working on this area, proposing definitions and objectives for \ac{xai}. Among the most relevant definitions, Gunning in 2017 describes \ac{xai} as \textit{“XAI will create a suite of \ac{ml} techniques that enables human users to understand, appropriately trust and effectively manage the emerging generation of artificially intelligent partner”} \cite{gunning2019xai}. Arrieta in 2019 defined \ac{xai} as \textit{“Given an audience, an explainable Artificial Intelligence is one that produces details or reasons to make its functioning clear or easy to understand.”} \cite{arrieta2020explainable}. From these definitions, it can be observed that the main objective of \ac{xai} is, on the one hand, creating a set of \ac{ml} techniques that produce more explainable models while maintaining an elevated level of performance. On the other hand, it also aims to allow humans to comprehend, properly trust, and efficiently manage the emergent generation of \ac{ai} partners \cite{gunning2019xai}. 

To make \ac{ai} models comprehensible to humans, two primary categories of models are distinguished: 

\begin{itemize}
    \item \textit{\textbf{Black-box models}} are characterized by their lack of interpretability, often due to their complexity or opaque nature. These models, such as deep neural networks, operate as "black boxes", where the internal decision-making processes are not easily accessible or understandable to humans \cite{molnar2020interpretable}. 
    \item \textit{\textbf{White-box models}} are inherently transparent and interpretable due to their simple and accessible structure. These allow users to trace and understand the logic behind their predictions without requiring additional tools or techniques \cite{lipton2018mythos}. 
\end{itemize}

To distinguish between these two types of models, one of the objectives \ac{xai} has, is to group the terms that make an \ac{ai} system or model clear and easy to understand. These terms had been independently studied, and \ac{xai} grouped them to define a model as a \textit{white-box} or a \textit{black-box} model: 

\begin{itemize}
    \item \textbf{Interpretability} can be defined as the ability to explain the meaning in understandable terms to a human \cite{miller2019explanation} when talking about AI.
    \item \textbf{Explainability} is the level to which a system can provide clarification about the cause of its decisions or results \cite{tomsett2018interpretable}. 
    \item \textbf{Transparency} is defined as the degree of understanding a model has by itself \cite{lipton2018mythos}. 
    \item \textbf{Understandability} is an equivalent term to intelligibility, which denotes the characteristic of the working of a model to be understood without the need for explaining the internal structure or the algorithmic means to process the data \cite{montavon2018methods}. 
    \item \textbf{Comprehensibility} is the term used to describe the ability of an algorithm to represent its knowledge in a human-understandable way \cite{cheng2022uncertainty}.  
\end{itemize}

Another goal \ac{xai} pursues, related to the previous ones, is to make the explanations generated by the \ac{ai} models dependent on the profile of the user receiving them \cite{arrieta2020explainable}. For example, high-knowledge users may demand explanations at a lower level, allowing them to enhance their knowledge. In contrast, system and model developers may demand explanations focused on understanding the internal workings of the model and its performance. This is the reason why \ac{xai} pursues to reach different profiles and make understandable and accessible the \ac{ai} systems and models depending on the needs of the user.   

Therefore, based on this section, the objectives of \ac{xai} can be defined as: 1) creating a set of techniques that allow \ac{ai} systems and models to be understood in a clear and simple manner, 2) grouping the terms related to the understanding of the workings of \ac{ai} systems and models, and 3) adapting the explanations to the profile of the user that is attempting to understand a given model. 

\section{Properties of \ac{ai} models in \ac{xai}}\label{section3}
In \ac{xai}, specific properties must be evaluated to determine whether an \ac{ai} model is comprehensible and explainable to users. Authors and professionals usually consider the following classification to know the degree of understanding and explainability of an \ac{ai} model \cite{arrieta2020explainable}: 

\begin{itemize}
    \item \textbf{Trustworthiness} is considered as the confidence of whether a model will act as intended when facing a given problem. 
    \item \textbf{Causality.} \ac{xai} aims to identify potential causal relationships among data variables. 
    \item \textbf{Transferability.} Models are always bounded by constraints that should allow for their seamless transferability. 
    \item \textbf{Informativeness.} Explainable \ac{ml} models should give information about the problem being tackled. 
    \item \textbf{Confidence} should always be assessed on a model in which reliability is expected. 
    \item \textbf{Fairness.} Explainability can be considered as the capacity to reach and guarantee fairness.
    \item \textbf{Accessibility.} Explainability is sometimes defined as a property that allows end users to get more involved in the process of improving and developing a certain \ac{ml} model. 
    \item \textbf{Interactivity.} The ability of a model to be interactive with the user is one of the goals targeted by an explainable \ac{ml} model. 
    \item \textbf{Privacy awareness} is an ability enabled by explainability in \ac{ml} models that may have complex representations of the learned patterns.  
\end{itemize}

These properties let us know the degree of understandability and explainability an \ac{ai} system or model has. However, it is to be considered the fulfillment of these properties in the context the model is deployed. As mentioned earlier, in some critical contexts, such as aeronautics and aerospace, where an elevated level of precision is desired to avoid harm to human life, damage to the environment, or significant economic losses, some properties may not be satisfied \cite{sutthithatip2022explainable}. 

For this reason, when assessing the properties of the models, the performance-interpretability trade-off of the model must always be considered \cite{ran2019survey}. All properties must be taken into account, always being conscious of them possibly compromising the precision of the system or model, making the deployment of it impossible in a critical environment. 

\section{XAI techniques}\label{section4}
\begin{figure}
    \centering
    \includegraphics[width=0.75\linewidth]{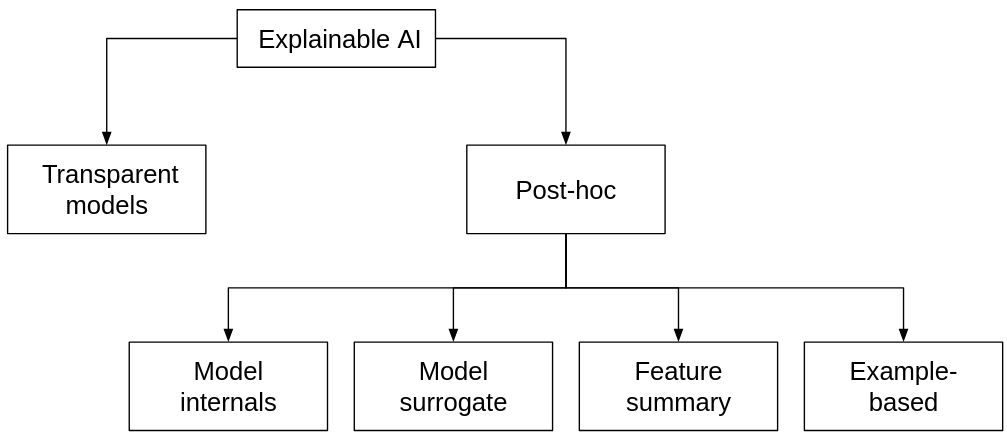}
    \caption{Taxonomy of \ac{xai} methods.}
    \label{fig:figure1}
\end{figure}

Both interpretable models and techniques to make opaque \ac{ai} systems and models comprehensible are included in \ac{xai} by its definition. As shown in figure \ref{fig:figure1}, a distinction is drawn between \textit{transparent models}, that are understandable given the simplicity of the algorithms, and the \textit{post-hoc techniques} applied after the model’s training to better understand its workings and predictions. 

A distinction is drawn between transparent models, which are inherently interpretable due to their simplicity, and post-hoc techniques, applied after model training to elucidate the internal workings and predictions of \textit{black-box} models. Furthermore, when analyzing them, it is to be considered that this kind of models can be tackled from the domain in which they are interpretable, that is \cite{arrieta2020explainable}: 

\begin{itemize}
    \item \textbf{Algorithmic transparency }deals with the ability of the user to understand the process followed by the model to produce any given output from the input data.
    \item \textbf{Decomposability} refers to a model’s ability to be understandable in smaller parts. This also requires each input to be interpretable. 
    \item \textbf{Simulatability} refers to the ability of a model to be manually simulated or conceptualized by a human user. 
\end{itemize}

Some examples among the models considered to be transparent are logistic and linear regressions given the straightforwardness to compute a prediction and the easiness to be understood by humans; decision trees, whose structure allow to give clear explanations by means of the visualization they provide; rule-based methods which are composed of a set of rules that make the comprehension of the predictions simple to humans; and additive models, such as GAM or GLM, that are extensions of regression models and treat each used variable to predict independently \cite{molnar2020interpretable}. 

Post-hoc techniques are applied to black-box models to be comprehensible by human beings. As seen in figure \ref{fig:figure1}, depending on the generated results, post-hoc techniques are classified into four different approaches \cite{molnar2020interpretable}:  

\begin{itemize}
    \item \textbf{Model internals.} Internal components and mechanisms considered to be interpretable are included. This includes components such as the weights in linear models, showing the relevance and direction of the relationship between characteristics; or the divisions that can be done in decision trees, that show how decisions are made based on sequential rules.
    \item \textbf{Model surrogate.} Techniques that approximate a model in a local (a part or instance of a model) or global (of the whole model) manner by means of models considered to be transparent. Within these are included LIME, which provides comprehensible explanations of black-box models by a local approximation of the predictions; Anchors, that explains the predictions of a model with highly precise local-specific rules; or SHAP, which obtains both local explanations by decomposing the prediction of an instance to individual contributions of the characteristics, and global explanations by summarizing the contributions for multiple instances. 
    \item \textbf{Feature summary.} This includes techniques that yield numerical statistics extracted after the model’s training. This includes classical summary statistics, feature importance (which allows for observing the weight of each feature in a prediction), or partial dependence plots (PDP) that help understand how one or more specific features influence a model’s prediction. 
    \item \textbf{Example-based.} It encompasses the techniques used to provide individual explanations for data instances based on other data instances, which can be real or simulated. Among these are counterfactuals, which analyze which changes must be done to an instance to obtain a different prediction; Influential Observations, which examine the variation of the prediction of a model when influential data instances are removed; and Prototypes and Criticisms, which show how average and most unusual examples look like for the user to better understand the data. 
\end{itemize}

\section{Applications of \ac{xai} in aeronautics and aerospace}\label{section5}
The need to understand how decisions are made within \ac{ai} systems and models has led to an increase in the use of \ac{xai} in aeronautics and aerospace. This need is particularly relevant in these fields because, as mentioned in Section 3, they are considered critical environments, and decisions made by \ac{ai} systems and models can have severe consequences. Therefore, developers must exercise exceptional care when designing these systems, ensuring they achieve a high degree of accuracy while being still interpretable and understandable by the humans who use them. In other words, they must consider the trade-off between accuracy and interpretability \cite{sutthithatip2022explainable}. This section presents various use cases where \ac{xai} has been applied in aeronautics and aerospace. 

In the aeronautical sector, applications of \ac{xai} can be found in various areas. In the most critical area of aeronautics, Air Traffic Management (ATM), the use of \ac{xai} has a significant impact on predictive tasks such as takeoff and landing times or incident risk assessment. Considerable efforts are made to explain the predictions generated by models based on neural networks \cite{degas2022survey}. Another application where \ac{xai} is having a significant impact is in the design and development of Unmanned Aerial Vehicles (UAVs) and drones. \ac{xai} is used in route adaptation during missions under adverse conditions, using fuzzy rules to explain how routes are adjusted during the missions these vehicles carry out \cite{keneni2019evolving}. It is also applied in simulations in three different modes to assess drone operations, validating decisions using metrics such as Mean Square Error (MSE), Root Mean Square Error (RMSE), and Mean Absolute Error (MAE) \cite{mualla2022quest}. Finally, another case is in post-natural disaster damage assessment, where data collected from drones and satellites is used to evaluate the damage caused by natural disasters, and explanations are generated using actual and predicted values, facilitating decision-making in emergency scenarios \cite{banimelhem2023explainable}. 

In aerospace, \ac{xai} applications are also found in various areas such as predictive maintenance. By applying post-hoc techniques like LIME or SHAP to DNNs used for the health management of vehicles integrated into aircrafts, the functioning of these neural networks was explained locally in aspects such as predictive accuracy, stability, and consistency of the models \cite{shukla2020opportunities}. On the other hand, in anomaly detection in spacecraft telemetry, the LIME technique was applied by analyzing different data instances used to train various models, which were then explained and exemplified for each type of anomaly in which the spacecraft telemetry signals were classified \cite{cuellar2024explainable}. Finally, within the area of satellite image processing, the use of decision trees combined with object detection through deep networks can be found for predicting poverty indices in Uganda, where the importance of features was used to identify which, visual elements had the greatest impact on the predictions \cite{banimelhem2023explainable}. Another application involves the use of attribution maps, such as Grad-CAM, to analyze daytime satellite images and nighttime light data in Sub-Saharan Africa \cite{ayush2021efficient}. 

\section{Conclusions}\label{section6}
In this paper, the presentation of \ac{xai} has been carried out, which can be defined as a set of \ac{ml} techniques that enable humans to understand and trust artificial intelligence systems and models. The objectives pursued by \ac{xai} have also been defined, which are; first, to create a set of techniques that allow artificial intelligence systems and models to be understood by humans; second, to group the various terms that may be related to the functioning of these models and systems; and third, to adapt the explanations and conclusions drawn based on the profile of the humans to whom they are directed. 

On the other hand, the properties that should be evaluated in \ac{ai} systems and models to be considered black-box models have been defined. Additionally, the concept of trade-off has been introduced, which allows developers of \ac{ai} systems and models to consider which properties need to be met in the environment where the systems and models are being developed to ensure a high degree of accuracy and a high degree of model interpretability. 

Additionally, the distinction has been made between white-box models, which are considered interpretable due to their algorithmic simplicity, and black-box models, which are not considered interpretable. Achieving transparency in black-box models requires the application of post-hoc techniques, which enable model interpretability and can be grouped into the following categories: Model Internals, which includes the internal components and mechanisms of models that are considered interpretable within transparent models; Model surrogate, which encompasses techniques that approximate a model locally or globally using models deemed transparent; Feature summary, which includes techniques that generate numerical statistics extracted after model training and during prediction, along with visualizations of these statistics; and Example-based, which includes techniques used to provide individual explanations of data instances using other data instances, whether real or simulated. 

Finally, various applications of \ac{xai} in aeronautics and aerospace have been presented. These sectors are considered critical environments, requiring that the systems and models developed ensure a high degree of accuracy alongside a high degree of interpretability. On one hand, it has been proven how the use of post-hoc techniques can ensure the interpretability of models in areas such as ATM or the design and development of drones in the aeronautical sector. On the other hand, in aerospace, these techniques can be applied to systems and models for predictive maintenance of spacecraft, anomaly detection in spacecraft telemetry, or satellite image processing. 

\section*{Acknowledgments}
This work has been carried out within the framework of the “Cátedra ENIA IA3: Cátedra de Inteligencia Artificial en Aeronáutica y Aeroespacio", subsidized by the “Ministerio de Asuntos Económicos y Transformación Digital (Secretaría de Estado de Digitalización e Inteligencia Artificial), del Gobierno de España”.

\newpage

\bibliography{references}

\end{document}